\pdfoutput=1
\documentclass[10pt,twocolumn,letterpaper]{article}
\usepackage[pagenumbers]{cvpr}
\usepackage{amsmath,amssymb,booktabs,array,graphicx,placeins}
\usepackage{tikz}
\usepackage{fontawesome5}
\usepackage{colortbl}
\usetikzlibrary{arrows.meta,positioning,fit,calc,backgrounds,patterns}
\definecolor{cvprblue}{rgb}{0.21,0.49,0.74}
\usepackage[breaklinks,colorlinks,allcolors=cvprblue]{hyperref}
\hypersetup{pdftitle={WM-Bench: A Diagnostic Benchmark for Working Memory in Egocentric Visual Assistants},pdfauthor={Amir Mousavi, Leslie Neely, John Davis, John Quarles}}
\input{glyphtounicode}\pdfgentounicode=1
\def\confName{CVPR}
\def\confYear{2027}

\newcommand{\ArcGainSmallR}{+0.205 [+0.172, +0.239]}

\newcommand{\ArcOffSmallR}{0.170}

\newcommand{\ArcOnSmallR}{0.375}

\newcommand{\ArcRecoveredR}{0.346}
\newcommand{\ArcRetrieveK}{seven}

\newcommand{\AuditPrecision}{0.805 [0.747, 0.860]}

\newcommand{\CapLvlGatedLruSmallR}{0.369}

\newcommand{\CapLvlLruSmallR}{0.210}

\newcommand{\CommonPairs}{81{,}995}

\newcommand{\EpEOneComplete}{$-$0.139 [$-$0.181, $-$0.099]}

\newcommand{\EpROneComplete}{$-$0.084 [$-$0.126, $-$0.045]}

\newcommand{\FamilyAlpha}{0.05}

\newcommand{\FramesPerSecond}{four}
\newcommand{\GapEwMinusFourteenFifoR}{+0.056 [+0.035, +0.077]}

\newcommand{\GateInterR}{+0.294 [+0.224, +0.366]}

\newcommand{\GateSmallR}{+0.159 [+0.132, +0.185]}

\newcommand{\HandDcDefaultC}{0.696}
\newcommand{\HandDcHandledC}{0.639}

\newcommand{\HandEpsDefaultC}{8.77}

\newcommand{\HandEpsHandledC}{2.13}

\newcommand{\HandEteaDefaultSmall}{0.149}

\newcommand{\HandEteaHandledSmall}{0.297}

\newcommand{\HandGainSmall}{+0.245 [+0.197, +0.295]}
\newcommand{\HandSources}{forty-nine}

\newcommand{\HeldGroups}{eighty}

\newcommand{\HoursTotal}{11.55}

\newcommand{\ItemLimitLarge}{192}
\newcommand{\ItemLimitMax}{448}
\newcommand{\ItemLimitMid}{sixty-four}
\newcommand{\ItemLimitSmall}{seven}

\newcommand{\ItemLimitTwenty}{twenty}

\newcommand{\JointLevel}{99.81\%}

\newcommand{\MarginPrimary}{0.02}

\newcommand{\MergeWindowS}{two}

\newcommand{\MolmoSources}{fifty-one}

\newcommand{\NumHypotheses}{twenty-seven}
\newcommand{\NumPairs}{123{,}472}
\newcommand{\NumPanelPolicies}{sixteen}

\newcommand{\NumSources}{101}

\newcommand{\OrdSplitSmallR}{0.88}

\newcommand{\OrdTauSmallR}{0.18}

\newcommand{\PoolSize}{eighteen}
\newcommand{\ProbeTickS}{one}
\newcommand{\ProxyCount}{four}

\newcommand{\RdPadMaxHit}{0.085}

\newcommand{\RdPadMaxMinusOracle}{$-$0.303 [$-$0.384, $-$0.223]}

\newcommand{\STwoFifoNncf}{0.030}

\newcommand{\STwoForeignHitThree}{0.041}

\newcommand{\STwoGatedLrfuNncf}{0.095}

\newcommand{\STwoLOne}{+0.347 [+0.266, +0.429]}

\newcommand{\STwoLTwo}{+0.060 [+0.025, +0.097]}

\newcommand{\STwoOracleHitThree}{0.388}

\newcommand{\STwoProbes}{7{,}900}
\newcommand{\STwoProbesNncf}{3{,}536}

\newcommand{\STwoWeightedNncf}{0.090}

\newcommand{\SepPairsSmall}{1431}

\newcommand{\SepPoliciesSmall}{54}

\newcommand{\SepSepSmallR}{68\%}

\newcommand{\SepTauSmallR}{0.87}

\newcommand{\SigAllControlsFail}{four}

\newcommand{\SigPassPanelCap}{Eleven}

\newcommand{\StageHeldQwen}{12.3\%}

\newcommand{\StageMissQwen}{11.0\%}

\newcommand{\StageNoPriorQwen}{9.1\%}

\newcommand{\StageNotHeldQwen}{67.6\%}

\newcommand{\TabOneProxyEmLlmMatched}{LRU}

\newcommand{\TabOneProxyGenAgentsMatched}{LRFU, 30\,s}

\newcommand{\TabOneProxyMemoryBankMatched}{LRFU, 15\,s}

\newcommand{\TabOneProxyMovieChatMatched}{Deduplicated FIFO}

\newcommand{\TargetActions}{five}
\newcommand{\TargetCap}{five}
\newcommand{\TargetHorizonS}{thirty}

\newcommand{\TauVlmMolmo}{0.84}
\newcommand{\TauVlmMolmoPolicies}{thirty-four}

\newcommand{\WmFourteenNames}{fourteen}

\providecolor{wmBlue}{HTML}{1F5FAE}\providecolor{wmOrange}{HTML}{C0661F}\providecolor{wmPurple}{HTML}{7E57C2}
\providecolor{wmInk}{HTML}{1F2933}\providecolor{wmGray}{HTML}{6B7785}\providecolor{wmSlate}{HTML}{4A5868}
\providecommand{\stg}[3]{\tikz[baseline=-.62ex]{\node[circle,fill=#1,text=white,minimum size=.36cm,inner sep=0,font=\fontsize{6.3}{6.5}\selectfont] {#2};}\,#3}
\providecommand{\grp}[3]{\tikz[baseline=-.62ex]{\node[circle,fill=#1,text=white,minimum size=.36cm,inner sep=0,font=\fontsize{6.3}{6.5}\selectfont] {#2};}\,\emph{#3}}
\providecommand{\cachetag}[1]{\tikz[baseline=-.62ex]{\node[rounded corners=3.5pt,fill=wmGray!10,draw=wmGray!55,line width=.45pt,inner xsep=2.6pt,inner ysep=1.3pt,font=\fontsize{7.4}{8}\selectfont,text=wmInk] {#1};}}
\providecommand{\anReg}{\tikz[baseline=-.62ex]{\node[rounded corners=3.5pt,fill=wmInk,inner xsep=2.6pt,inner ysep=1.3pt,font=\fontsize{6.6}{7}\selectfont,text=white] {Registered};}}
\providecommand{\anPost}{\tikz[baseline=-.62ex]{\node[rounded corners=3.5pt,draw=wmGray,line width=.45pt,inner xsep=2.6pt,inner ysep=1.3pt,font=\fontsize{6.6}{7}\selectfont,text=wmInk] {Post-reg.};}}
\providecommand{\anSec}{\tikz[baseline=-.62ex]{\node[rounded corners=3.5pt,fill=wmGray!15,draw=wmGray,line width=.45pt,inner xsep=2.6pt,inner ysep=1.3pt,font=\fontsize{6.6}{7}\selectfont,text=wmInk] {Secondary};}}
\providecommand{\anCtl}{\tikz[baseline=-.62ex]{\node[rounded corners=3.5pt,draw=wmGray!60,line width=.45pt,dash pattern=on 1.5pt off 1pt,inner xsep=2.6pt,inner ysep=1.3pt,font=\fontsize{6.6}{7}\selectfont,text=wmGray] {Control};}}
\providecommand{\fpZero}{\draw[wmGray!70,line width=.4pt,dash pattern=on 1.2pt off 1.2pt] (0.725,-.14) -- (0.725,.24);}
\newcommand{\fpHandGainSmall}{\tikz[baseline=-.6ex,x=1cm,y=1cm]{\useasboundingbox (0,-.14) rectangle (1.45,.24);\fpZero\draw[wmBlue,line width=1.6pt,line cap=round] (1.042,.05) -- (1.200,.05);\fill[wmBlue] (1.120,.05) circle (.075);}}
\newcommand{\fpGateSmallR}{\tikz[baseline=-.6ex,x=1cm,y=1cm]{\useasboundingbox (0,-.14) rectangle (1.45,.24);\fpZero\draw[wmPurple,line width=1.6pt,line cap=round] (0.938,.05) -- (1.023,.05);\fill[wmPurple] (0.981,.05) circle (.075);}}
\newcommand{\fpGateInterR}{\tikz[baseline=-.6ex,x=1cm,y=1cm]{\useasboundingbox (0,-.14) rectangle (1.45,.24);\fpZero\draw[wmPurple,line width=1.6pt,line cap=round] (1.086,.05) -- (1.315,.05);\fill[wmPurple] (1.199,.05) circle (.075);}}
\newcommand{\fpArcGainSmallR}{\tikz[baseline=-.6ex,x=1cm,y=1cm]{\useasboundingbox (0,-.14) rectangle (1.45,.24);\fpZero\draw[wmOrange,line width=1.6pt,line cap=round] (1.002,.05) -- (1.110,.05);\fill[wmOrange] (1.055,.05) circle (.075);}}
\newcommand{\fpSTwoLOne}{\tikz[baseline=-.6ex,x=1cm,y=1cm]{\useasboundingbox (0,-.14) rectangle (1.45,.24);\fpZero\draw[wmSlate,line width=1.6pt,line cap=round] (1.154,.05) -- (1.416,.05);\fill[wmSlate] (1.284,.05) circle (.075);}}
\newcommand{\fpSTwoLTwo}{\tikz[baseline=-.6ex,x=1cm,y=1cm]{\useasboundingbox (0,-.14) rectangle (1.45,.24);\fpZero\draw[wmSlate,line width=1.6pt,line cap=round] (0.765,.05) -- (0.881,.05);\fill[wmSlate] (0.822,.05) circle (.075);}}
\newcommand{\fpRdPadMaxMinusOracle}{\tikz[baseline=-.6ex,x=1cm,y=1cm]{\useasboundingbox (0,-.14) rectangle (1.45,.24);\fpZero\draw[wmSlate,line width=1.6pt,line cap=round] (0.106,.05) -- (0.366,.05);\fill[wmSlate] (0.237,.05) circle (.075);}}
\newcommand{\fpEpEOneComplete}{\tikz[baseline=-.6ex,x=1cm,y=1cm]{\useasboundingbox (0,-.14) rectangle (1.45,.24);\fpZero\draw[wmPurple,line width=1.6pt,line cap=round] (0.433,.05) -- (0.565,.05);\fill[wmPurple] (0.501,.05) circle (.075);}}
\newcommand{\fpEpROneComplete}{\tikz[baseline=-.6ex,x=1cm,y=1cm]{\useasboundingbox (0,-.14) rectangle (1.45,.24);\fpZero\draw[wmPurple,line width=1.6pt,line cap=round] (0.522,.05) -- (0.652,.05);\fill[wmPurple] (0.590,.05) circle (.075);}}
\newcommand{\fpAxis}{\tikz[baseline=0pt,x=1cm,y=1cm]{\useasboundingbox (0,0) rectangle (1.45,.32);\draw[wmGray,line width=.4pt] (0,0) -- (1.45,0);\draw[wmGray,line width=.4pt] (0.081,0) -- (0.081,.06) node[anchor=south west,inner sep=.8pt,font=\fontsize{6.5}{7}\selectfont,text=wmGray] {$-$0.4};\draw[wmGray,line width=.4pt] (1.369,0) -- (1.369,.06) node[anchor=south east,inner sep=.8pt,font=\fontsize{6.5}{7}\selectfont,text=wmGray] {$+$0.4};\draw[wmGray,line width=.4pt] (0.725,0) -- (0.725,.06);}}
\newcommand{\RegRSeven}{+0.267 [+0.191, +0.345]}
\newcommand{\fpRegRSeven}{\tikz[baseline=-.6ex,x=1cm,y=1cm]{\useasboundingbox (0,-.14) rectangle (1.45,.24);\fpZero\draw[wmGray,line width=1.6pt,line cap=round] (1.032,.05) -- (1.280,.05);\fill[wmGray] (1.155,.05) circle (.075);}}

\let\wmincludegraphics\includegraphics

\title{WM-Bench: A Diagnostic Benchmark for Working Memory in Egocentric Visual Assistants}
\author{%
Amir Mousavi\textsuperscript{1}\qquad
Leslie Neely\textsuperscript{2}\qquad
John Davis\textsuperscript{3}\qquad
John Quarles\textsuperscript{1}\\[0.6ex]
{\normalsize \textsuperscript{1}Department of Computer Science, College of AI, Cyber and Computing}\\
{\normalsize \textsuperscript{2}Department of Neuroscience, Developmental and Regenerative Biology, College of Sciences}\\
{\normalsize \textsuperscript{3}Department of Educational Psychology, College of Education and Human Development}\\
{\normalsize The University of Texas at San Antonio}\\[0.4ex]
{\ttfamily\small \{seyedamir.mousavi, leslie.neely, john.davis2, john.quarles\}@utsa.edu}
}
\begin{document}
\maketitle
\begin{abstract}
Egocentric visual assistants must keep information about previously observed objects accessible while new observations arrive under limited working-memory capacity. An object's relevance may become clear only during a later action. A vision--language model (VLM) reports object names; working memory retains some, long-term memory retrieves older names, and a reader uses the current frame and available names to predict the next object. End-to-end accuracy cannot identify where needed information was lost.

We introduce \textbf{WM-Bench}, a benchmark that separates these stages using fixed VLM name streams. It evaluates \SepPoliciesSmall{} memory policies on \NumSources{} held-out egocentric recordings (\HoursTotal{} hours), reserving future-action annotations for independent scoring and offline reference policies. Policies include conventional caches, cognitively inspired implementations, and proxies for published memory systems, with separate behavioral tests.

With seven stored names, selecting the first reported name per frame improves retention by 15.9 percentage points over the same least-recently-used cache without selection; retrieval raises availability by 20.5 points. Where the current frame's name report omits the target, a list built from future actions improves next-object prediction within three guesses by 34.7 points over an unrelated list. A deployable policy combining selective admission with recency and frequency improves by 6.0 points over first-in-first-out memory. WM-Bench distinguishes what is reported, retained, retrieved, and used. The pre-registration, CPU-replay code, cached name streams, audit materials, and reproduction instructions will be released upon acceptance.
\end{abstract}

\section{Introduction}
\label{sec:intro}
\begin{figure*}[t]
\centering
\wmincludegraphics[width=0.97\textwidth]{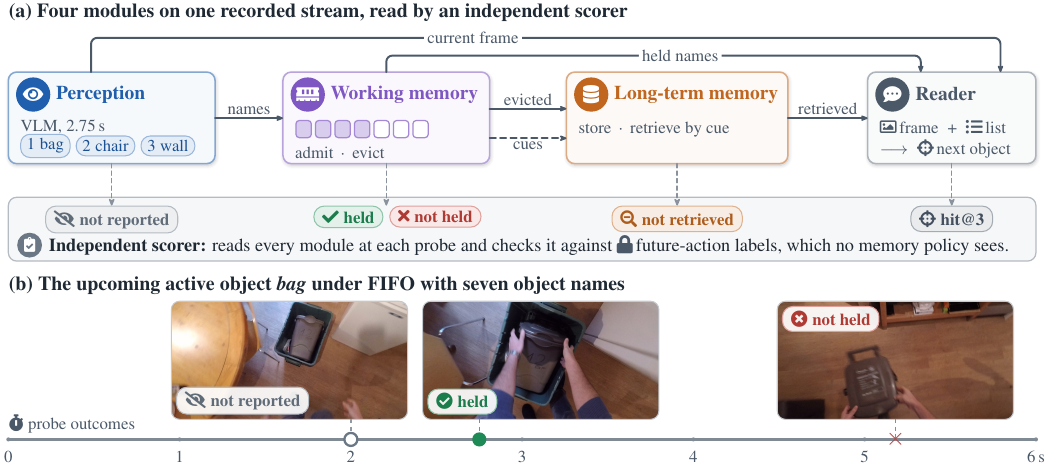}
\caption{\textbf{WM-Bench follows object-name availability across stages.} (a) Perception sends names to bounded working memory; discarded names enter a long-term archive. The reader uses the current frame, held names, and retrieved names. An independent scorer receives future-action labels, never passed to deployable policies. (b) Original EPIC-KITCHENS-100 frames (P01\_14) with a seven-name first-in-first-out (FIFO) memory show \emph{bag} unreported (2.00\,s), held (2.75\,s), then no longer held (5.18\,s; frame at 5.25\,s). The markers are probe outcomes, not an exact eviction time.}
\label{fig:teaser}
\end{figure*}

Egocentric visual assistants may need objects long after they leave the camera's view: a cook may set aside a lid and need it several actions later~\cite{grauman2022ego4d,hodges2006sensecam,berry2007sensecam}. If the assistant never names the lid, forgets it, or buries it in a long list, it may fail to suggest it at the right step. Vision--language models (VLMs) can name objects in successive frames, but keeping every name can distract the model asked to predict the next object~\cite{shi2023distracted,levy2024same,liu2024lost}. We therefore study a \emph{bounded, online working memory} that keeps a small set of reported names. We distinguish this store from \emph{perception} (which reports object names), \emph{long-term memory} (which archives and retrieves older names), and the \emph{reader} (a model that uses the current frame and available names; Fig.~\ref{fig:teaser}).

Deciding what to retain is difficult because the usefulness of an observed object may become clear only when the wearer acts later. Before then, a system must decide \emph{which names to request}, \emph{which to admit}, and \emph{which to evict}; later, it must decide \emph{what to retrieve} and \emph{what to show the reader}. Video and agent memories draw on cognitive notions of decay, consolidation, and importance~\cite{cheng2022xmem,song2024moviechat,zhong2024memorybank,park2023generative,fountas2025emllm}. Yet cognitive theories often leave implementation choices open~\cite{guest2021how,cooper2014implementations}, and conventional cache policies already keep items by how recently or how often they appeared~\cite{lee2001lrfu,megiddo2003arc}. Thus, strong retention does not establish that a borrowed cognitive operation caused the gain. Likewise, final-answer accuracy can conflate an object never named, a name discarded from memory, a failed retrieval, and a reader that overlooks an available name.

\textbf{WM-Bench asks which stages and decisions make previously observed objects accessible and useful under a strict streaming-memory budget.} Answering this requires separating upstream name delivery from bounded retention, distinguishing held names from retrieved names, and testing whether a reader actually benefits from the result. It also requires asking what a cognitively motivated rule adds beyond a conventional cache using the same information. These are connected questions about one assistant pipeline, not separate benchmarks.

We introduce \textbf{WM-Bench} to connect these tests. It generates object-name streams once from egocentric recordings and replays the same timestamped events causally through each memory policy at an identical item-count limit. Later action annotations define which object classes will soon be used, but never inform a deployable policy. At regular probes, upcoming active objects are defined by the next \TargetActions{} annotated actions beginning within \TargetHorizonS{} seconds. Future annotations are reserved for scoring and for reference policies that can look ahead (oracles), never for deployable policies. WM-Bench distinguishes \emph{Delivery Coverage} (was a needed class named?), \emph{Retention after Delivery} (was a delivered class kept?), and \emph{End-to-End Availability} (is a needed class currently held?). Long-term retrieval and whether the reader identifies the next object among its first three guesses are evaluated separately. Consequently, a perception omission is not mistaken for a replacement failure, and an available name is not automatically counted as a correct answer (Sec.~\ref{sec:design}).

Our evaluation covers \SepPoliciesSmall{} policies and \NumSources{} held-out recordings from Ego4D, EPIC-KITCHENS-100, and HD-EPIC (\HoursTotal{} hours)~\cite{grauman2022ego4d,damen2022rescaling,perrett2025hdepic}. It combines a registered analysis plan, written policy specifications, controlled replay, tests of predicted cognitive behavior, caches with the same ranking information, and next-object reader experiments. The principal findings form a continuous account: more selective delivery and admission improve small-memory retention; long-term retrieval restores older names; the reader benefits from relevant short lists but is sensitive to distractors; and capacity and within-frame order change the interpretation of memory-policy comparisons (Sec.~\ref{sec:experiments}).

\paragraph{Contributions.}
We make three contributions:
\begin{itemize}
\item \textbf{A benchmark that identifies where a needed object becomes unavailable.} Under fixed perceptual streams and future-action probes, WM-Bench measures name delivery, bounded retention, long-term retrieval, and reader use separately; misses can be assigned to the stage where availability was lost.
\item \textbf{A controlled evaluation of \SepPoliciesSmall{} memory policies.} A common causal interface and held-out protocol connect cache baselines, cognitive and neuroscience models, and video/agent-memory proxies. Written specifications, behavioral tests, and caches using the same ranking information reveal what each implementation adds.
\item \textbf{Evidence for practical design choices.} Registered and separately identified additional analyses test admission, retrieval, capacity, list composition, perception instructions, and event order; the benchmark exposes which changes actually make needed objects accessible to a reader.
\end{itemize}

\section{Related Work}
\label{sec:related}
\paragraph{Egocentric and streaming-memory evaluation.}
Egocentric datasets and benchmarks—including Ego4D, EPIC-KITCHENS-100, and HD-EPIC—make it possible to associate earlier observations with later object interactions~\cite{grauman2022ego4d,damen2022rescaling,perrett2025hdepic}. Ego4D's natural-language queries and object-centric memories such as AMEGO and OSNOM support long-horizon localization or recovery of earlier observations~\cite{grauman2022ego4d,goletto2024amego,plizzari2025spatial}; their goal differs from deciding which \emph{few object names} a continuously updating working memory should hold now. StreamingBench, OVO-Bench, and VideoMME assess online or long-video understanding~\cite{lin2024streamingbench,niu2025ovobench,fu2025videomme}, while EgoStream, StreamMemBench, and M3Eval examine memory more directly~\cite{forte2026egostream,liu2026streammembench,huang2026m3eval}. SPACE and DeltaS address capacity-bounded video memories~\cite{niu2026space,kwon2026deltas}. LLM and agent-memory suites additionally test long-context retrieval and dialogue recall~\cite{hsieh2024ruler,maharana2024locomo,wu2025longmemeval}, while video needle tests locate sparse facts in long inputs~\cite{zhao2025videoniah}. StreamMemBench traces evidence use, and What Eviction Destroys isolates text-agent evictions; neither combines per-object visual delivery with fixed-capacity online retention~\cite{liu2026streammembench,shen2026eviction}. A final answer can still conflate a perception miss, eviction, retrieval failure, or reader behavior. \textbf{WM-Bench scores these stages separately on the same future-action targets while replaying identical reported-name streams across policies.} This is a focused symbolic-memory evaluation, not a replacement for full video understanding.

\paragraph{Controlled benchmarks and cache baselines.}
Controlled evaluations such as What Eviction Destroys, WritePolicyBench, FiFA, and Selective Memory Retention directly compare memory decisions~\cite{shen2026eviction,elcham2026writepolicybench,alqithami2025fifa,reddy2026traceretain}. Their controls motivate ours, and recent identical-stream comparisons also show the strength of simple memories~\cite{shen2026simplestream}. Retention alone, however, does not show whether an implementation behaves as its motivating theory predicts. We include first-in-first-out (FIFO), least-recently-used (LRU), least-recently/frequently-used (LRFU), least-frequently-used with dynamic aging (LFU-DA), ARC, S3-FIFO, and other caching policies as standard references~\cite{belady1966study,sleator1985amortized,lee2001lrfu,arlitt2000lfuda,megiddo2003arc,yang2023s3fifo}. For each cognitive or published-memory policy, we test a conventional cache that uses the same information, such as recency or frequency, but omits the policy's added operation. We call this its \emph{mechanism-matched cache}. Behavioral tests separately check whether the policy exhibits the effect predicted by the original study or model; passing does not prove better retention. Table~\ref{tab:benchcmp} compares these diagnostic capabilities across representative benchmarks; fuller comparisons are in the supplement.

\DeclareRobustCommand{\hbfull}{\tikz[baseline=-0.6ex]{\fill[black!75] (0,0) circle (0.62ex);\draw[black!75, line width=0.5pt] (0,0) circle (0.62ex);}}
\DeclareRobustCommand{\hbhalf}{\tikz[baseline=-0.6ex]{\draw[black!75, line width=0.5pt] (0,0) circle (0.62ex);\fill[black!75] (0,0.62ex) arc (90:270:0.62ex) -- cycle;}}
\DeclareRobustCommand{\hbnone}{\tikz[baseline=-0.6ex]{\draw[black!75, line width=0.5pt] (0,0) circle (0.62ex);}}
\DeclareRobustCommand{\hbna}{\textcolor{black!50}{--}}
\begin{table*}[t]
\centering\footnotesize
\setlength{\tabcolsep}{3pt}
\renewcommand{\arraystretch}{1.03}
\begin{tabular}{@{}>{\raggedright\arraybackslash}p{3.50cm}*{6}{>{\centering\arraybackslash}p{2.12cm}}@{}}
\toprule
Benchmark & \shortstack{Identical\\streams} & \shortstack{Held names\\scored directly} & \shortstack{Perception\\vs. memory} & \shortstack{Cache\\comparisons} & \shortstack{Same-signal\\cache control} & \shortstack{Predicted\\behavior tested} \\
\midrule
\multicolumn{7}{l}{\grp{wmBlue}{\faVideo}{Video and egocentric streaming benchmarks}} \\
SPACE~\cite{niu2026space}                          & \hbfull & \hbhalf & \hbhalf & \hbfull & \hbnone & \hbnone \\
EgoStream~\cite{forte2026egostream}                & \hbfull & \hbnone & \hbhalf & \hbfull & \hbnone & \hbnone \\
M3Eval~\cite{huang2026m3eval}                      & \hbnone & \hbnone & \hbhalf & \hbnone & \hbnone & \hbhalf \\
DeltaS~\cite{kwon2026deltas}                       & \hbfull & \hbnone & \hbnone & \hbfull & \hbnone & \hbnone \\
\midrule
\multicolumn{7}{l}{\grp{wmSlate}{\faRobot}{Agent-memory benchmarks}} \\
StreamMemBench~\cite{liu2026streammembench}        & \hbhalf & \hbhalf & \hbhalf & \hbnone & \hbnone & \hbnone \\
What Eviction Destroys~\cite{shen2026eviction}     & \hbfull & \hbfull & \hbhalf & \hbfull & \hbnone & \hbnone \\
WritePolicyBench~\cite{elcham2026writepolicybench} & \hbfull & \hbfull & \hbna   & \hbfull & \hbnone & \hbnone \\
FiFA~\cite{alqithami2025fifa}                      & \hbfull & \hbnone & \hbna   & \hbfull & \hbnone & \hbnone \\
Selective Memory Retention~\cite{reddy2026traceretain} & \hbfull & \hbhalf & \hbna & \hbfull & \hbnone & \hbnone \\
\midrule
\rowcolor{wmBlue!8}\textbf{WM-Bench (ours)}                         & \hbfull & \hbfull & \hbfull & \hbfull & \hbfull & \hbfull \\
\bottomrule
\end{tabular}
\caption{\textbf{What distinguishes WM-Bench from existing memory evaluations.} The six axes separate fixed-input policy testing, name-level scoring, stage diagnosis, cache baselines, caches using the same ranking information, and tests of predicted behavior. Filled circles mean full coverage, half-filled circles partial coverage, and open circles absent; a dash marks a stage not applicable to that benchmark.}
\label{tab:benchcmp}
\end{table*}

\paragraph{Cognitive, short-term, and long-term memories.}
Cognitive psychology and neuroscience study forgetting, consolidation, serial order, interference, and capacity~\cite{anderson1989human,ricker2014consolidation,page1998primacy,oberauerlin2024interference,bays2009precision}. In video and agent systems, XMem and MovieChat distinguish short- and long-term stores, MemoryBank adopts a forgetting curve, and Generative Agents use recency, importance, and relevance~\cite{cheng2022xmem,song2024moviechat,zhong2024memorybank,park2023generative}. EM-LLM segments memories at event boundaries~\cite{fountas2025emllm}; hierarchical streaming-video memories likewise separate fast and persistent stores~\cite{zhang2026hermes,xiong2025streamchat}. Such sources motivate algorithms but do not necessarily specify all event-by-event decisions in the passive VLM-name setting~\cite{guest2021how,cooper2014implementations}. WM-Bench makes those decisions explicit and tests them. Importantly, \emph{working-memory retention} (currently held names) differs from \emph{long-term retrieval} (archived names returned later); retrieved names do not occupy a seven-entry working-memory slot merely because they are accessible to the reader.

\paragraph{Retrieval and reader utilization.}
MemGPT and streaming language/video models manage the information exposed to a reader~\cite{packer2023memgpt,xiao2024streamingllm,qian2024streaming}. Long-context work shows distractor and position effects~\cite{liu2024lost,shi2023distracted,levy2024same}. WM-Bench holds the current frame and reader VLM fixed while changing only the bounded list of object names, testing whether the retained information is \emph{usable}, not merely present. Its reader task follows next-object anticipation in egocentric video~\cite{furnari2019rulstm,damen2022rescaling}.

\section{WM-Bench}
\label{sec:design}
\paragraph{One causal interface for the memory pipeline.}
\label{sec:scoring}\label{sec:status}
A VLM processes video frames and outputs timestamped \emph{object-name events}. Working memory receives each event in its stream order and updates its bounded state, $S_i=U(S_{i-1},e_i,\Delta t_i)$. The names reported in one frame have the same timestamp and enter in the VLM's listing order. A policy may admit, refresh, or evict a name, but cannot consult the future. Only the independent scorer and future-informed reference policies (oracles) see later action annotations. The reader receives the current frame and a list of names accessible from working memory, optionally augmented by retrieval (Fig.~\ref{fig:teaser}). We refer to a policy's state as \emph{held}, to an older name returned by long-term memory as \emph{retrieved}, and to their union as \emph{available after retrieval}; these are not the same outcome.

\paragraph{Recordings and two controlled streams.}
\label{sec:interface}
All choices were fixed using development recordings before the held-out evaluation. The held-out set contains \NumSources{} recordings in \HeldGroups{} independent recording groups (\HoursTotal{} hours) from Ego4D, EPIC-KITCHENS-100, and HD-EPIC~\cite{grauman2022ego4d,damen2022rescaling,perrett2025hdepic}. Qwen3-VL~\cite{qwen2025qwen3vl} names objects from \FramesPerSecond{} independently decoded frames per second. Its default instruction requests objects one could touch or pick up; a separate instruction requests active objects, those being handled or used. Names repeated within \MergeWindowS{} seconds are merged (the merge window); a fixed object-class table aligns reported names with later annotated uses (matching audits and stricter controls: supplement). Decoding and exclusion rules are fixed on development recordings and documented in the supplement.

The \emph{real stream} contains only VLM-generated names. The \emph{matched-pressure stream} additionally inserts an annotated object name only \emph{at the time of its observed action}, never before it occurs. It has similar event pressure (names per second) and tests replacement decisions with more complete delivery of reusable objects. The extra events do not represent a deployable VLM, and their action-time insertion must not be confused with future-target access by a memory policy. Policies are compared on identical input within each stream, and the real stream checks whether the effect persists without those additions. Both are distinct from the separately generated active-objects instruction stream.

\paragraph{Probes and stage-level measures.}
Probes occur every \ProbeTickS{} second, on every fourth sampled VLM frame, and at annotated action onsets. At a scoring probe, $G$ contains classes used in the next \TargetActions{} actions beginning within \TargetHorizonS{} seconds (at most \TargetCap{} classes); $O$ is the set of classes reported at or before the probe, $D=G\cap O$ the \emph{delivered} classes, and $B$ the classes \emph{held} now. A report's mere presence in a recording does not make a name held at a later probe. We measure
\begin{align}
\text{Delivery Coverage (DC)} &= |D|/|G|,\label{eq:dc}\\[-2pt]
\text{Retention after Delivery (RAD)} &= |B\cap D|/|D|,\label{eq:rad}\\[-2pt]
\text{End-to-End Availability (EtEA)} &= |B\cap G|/|G|.\label{eq:etea}
\end{align}
DC evaluates perception; RAD evaluates memory conditional on delivery; EtEA evaluates what working memory makes accessible among all needed classes. Retrieval availability scores $|(B\cup R)\cap G|/|G|$, where $R$ is the set returned by long-term memory, and is \emph{not} RAD. The reader is scored separately by whether the correct next object is among its three guesses (hit@3). Empty-denominator probes are excluded. Following the recurrent-object motivation of egocentric action studies~\cite{pirsiavash2012adl}, the primary population is \CommonPairs{} of \NumPairs{} probe--object pairs for classes used in an earlier annotated action. Probe-level ratios are averaged by recording and then equally across recording groups, which are the units of inference.

\paragraph{Policies, capacities, and mechanism controls.}
\label{sec:policies}\label{sec:proxies}\label{sec:contracts}
We evaluate \SepPoliciesSmall{} policies with \ItemLimitSmall{} name entries and capacity sweeps up to \ItemLimitMax{}. A capacity of seven means seven stored entries, each containing an object name; it does not guarantee seven distinct categories or the same byte use across implementations. The policies include conventional caches; \NumPanelPolicies{} rules adapted from cognitive and neuroscience accounts (the \emph{panel}); three canonical models, ACT-R (recency and frequency), SIMPLE (temporal distinctiveness), and TBRS (refreshing over time); \ProxyCount{} symbolic versions of published video/agent memories; and two future-informed references. The panel covers serial order, refreshing, interference, association, consolidation, and event boundaries. Written specification cards distinguish formal models (specified updates), quantitative principles (specified effects), and empirical findings (predicted behaviors). They record the missing implementation choices, constants, and unavailable inputs; none are fitted on held-out recordings.

A \emph{behavioral test} uses controlled sequences of names to check an effect predicted by its original study or model, then repeats the test without the proposed operation and with five control policies~\cite{roberts2000persuasive,oberauer2018benchmarks}. For example, Post-Arrival Consolidation predicts that a name followed by more event-free time remains accessible longer; its test varies that interval, and its ablation removes consolidation. Separately, a \emph{mechanism-matched cache} ranks names using the same information (such as recency or frequency), without the added cognitive operation. Comparing retention measures what that operation contributes beyond a conventional cache. Table~\ref{tab:policy_main} gives examples. The supplement provides the full policy-to-source matrix, source tests, per-policy comparisons, and implementation details. We also compare with the strongest eligible development-selected cache and future-informed reference policies, which are not deployable or guaranteed upper bounds. The eligible cache is selected only from development recordings among \PoolSize{} capacity-bounded replacement policies. The capacities are \ItemLimitSmall{}, \ItemLimitTwenty{}, \ItemLimitMid{}, \ItemLimitLarge{}, and \ItemLimitMax{} entries; occupancy, distinct names, byte-matched controls, and CPU costs are audited (supplement).

\begin{table*}[t]
\centering\footnotesize
\setlength{\tabcolsep}{5pt}
\renewcommand{\arraystretch}{1.07}
\begin{tabular}{@{}>{\raggedright\arraybackslash}p{4.1cm}>{\raggedright\arraybackslash}p{5.1cm}>{\raggedright\arraybackslash}p{3.05cm}>{\raggedright\arraybackslash}p{4.2cm}@{}}
\toprule
Policy (source) & Translated operation & Matched cache & Source or system signal \\
\midrule
\multicolumn{4}{@{}l}{\grp{wmPurple}{\faBrain}{Cognitive and neuroscience mechanisms}} \\
Post-Arrival Consolidation~\cite{ricker2014consolidation} & Strengthen an item after time free of events & \cachetag{LRFU (15 s)} & Free-time consolidation \\
Similarity Interference~\cite{oberauerlin2024interference} & Reduce the retention of competing names & \cachetag{LRU} & Similarity-based competition \\
Serial-Order Competition~\cite{page1998primacy} & Retain according to arrival position & \cachetag{FIFO} & Serial position \\
ACT-R base-level eviction~\cite{anderson1998atomic} & Evict according to base-level activation & \cachetag{LFU-DA} & Recency and frequency \\
TBRS time-based refreshing~\cite{oberauer2011tbrs} & Refresh names as temporal resources permit & \cachetag{LRFU (1 s)} & Time-based refreshing \\
\midrule
\multicolumn{4}{@{}l}{\grp{wmOrange}{\faFilm}{Published video and agent-memory proxies}} \\
MemoryBank~\cite{zhong2024memorybank} & Symbolic forgetting-curve eviction & \cachetag{\TabOneProxyMemoryBankMatched} & Forgetting curve \\
Generative Agents~\cite{park2023generative} & Rank names by source-inspired importance & \cachetag{\TabOneProxyGenAgentsMatched} & Recency / importance / relevance \\
EM-LLM~\cite{fountas2025emllm} & Event-boundary-based symbolic replacement & \cachetag{\TabOneProxyEmLlmMatched} & Event segmentation \\
MovieChat~\cite{song2024moviechat} & Symbolic short/long-term replacement & \cachetag{\TabOneProxyMovieChatMatched} & Two-store management \\
\bottomrule
\end{tabular}
\caption{\textbf{Representative translations under the same bounded-name interface.} These examples show how source mechanisms and published-system rules become causal eviction policies, and which cache uses the corresponding ranking information. The complete matrix of \NumPanelPolicies{} panel policies, canonical models, \ProxyCount{} proxies, synthetic behavior tests, and retention contrasts appears in the supplement. The proxies reproduce symbolic eviction, not entire multimodal systems.}
\label{tab:policy_main}
\end{table*}

\paragraph{Registered comparisons and statistical inference.}
Before inspecting the held-out results, we registered \NumHypotheses{} hypotheses, including which comparisons belong to the same statistical testing family. The planned comparisons cover mechanism-matched retention, admission across capacities, within-frame order, perception noise (omitted or false names), retrieval variants (associative against frequency retrieval), reader-list contrasts, and frame-rate controls (one frame per second). Differences are tested configuration minus baseline. Paired group differences use exact sign-flip intervals~\cite{pitman1937significance}; single-configuration levels use group bootstrap intervals. The registered graphical procedure controls family-wise error~\cite{bretz2009graphical} at $\alpha=\FamilyAlpha{}$ and reports \JointLevel{} simultaneous bounds for the confirmatory contrasts; the practical RAD margin is $\pm\MarginPrimary{}$. We report supported effects with their intervals and retain the complete decision matrix in the supplement. Registered tests follow the prospective plan; post-registration analyses were added later, even when they use held-out recordings (Table~\ref{tab:results}). Additional analyses conducted after registration are labeled explicitly and are not promoted to pre-registered claims.

\section{Results}
\label{sec:experiments}
\paragraph{Reading the results.}
The evaluation uses \NumSources{} held-out recordings and the fixed policy interface of Sec.~\ref{sec:design}. We follow the stages a needed object traverses: delivery and admission, bounded retention, retrieval, then reader use. Unless specified, contrasts refer to the real stream with \ItemLimitSmall{} entries. In FIFO's stage breakdown, \StageMissQwen{} of probe--object pairs lack a prior VLM report, \StageNotHeldQwen{} were delivered but are no longer held, and \StageHeldQwen{} remain held; \StageNoPriorQwen{} concern classes without an earlier annotated action. This makes the distinction between perception coverage and memory retention consequential. Table~\ref{tab:results} selects the principal, stage-specific quantitative comparisons; full policy curves, uncertainty intervals, and hypothesis-family decisions are in the supplement.

\begin{table*}[t]
\centering\footnotesize
\setlength{\tabcolsep}{2pt}
\renewcommand{\arraystretch}{1.18}
\begin{tabular}{@{}>{\raggedright\arraybackslash}p{2.85cm}>{\raggedright\arraybackslash}p{8.05cm}>{\raggedright\arraybackslash}p{3.25cm}c>{\raggedright\arraybackslash}p{1.2cm}@{}}
\toprule
Pipeline stage & Tested configuration $-$ reference (measure) & Difference [interval] & \fpAxis & Analysis \\
\midrule
\stg{wmBlue}{\faEye}{Perception} & Active-objects $-$ default instruction; FIFO, seven entries (RAD) & \HandGainSmall{} & \fpHandGainSmall & \anPost \\
\stg{wmPurple}{\faFilter}{Admission} & LRU with $-$ without first-listed admission, seven entries (RAD) & \GateSmallR{} & \fpGateSmallR & \anSec \\
 & Gain of first-listed admission at 7 minus 192 entries (RAD, LRU) & \GateInterR{} & \fpGateInterR & \anReg \\
\midrule
\stg{wmOrange}{\faDatabase}{Long-term retrieval} & FIFO with $-$ without archive retrieval (availability, real stream) & \ArcGainSmallR{} & \fpArcGainSmallR & \anSec \\
\midrule
\stg{wmSlate}{\faCommentDots}{Reader} & Future-informed reference list $-$ unrelated list (hit@3) & \STwoLOne{} & \fpSTwoLOne & \anReg \\
 & ACT-R rule with first-listed admission $-$ FIFO list (hit@3) & \STwoLTwo{} & \fpSTwoLTwo & \anReg \\
 & Future-informed list padded to 192 $-$ same seven-name list (hit@3) & \RdPadMaxMinusOracle{} & \fpRdPadMaxMinusOracle & \anReg \\
\midrule
\stg{wmGray}{\faRandom}{Order robustness} & Isolation-Weighted Decay $-$ FIFO: listing $-$ reversed order (RAD) & \RegRSeven{} & \fpRegRSeven & \anReg \\
 & Listing versus reversed within-frame order (Kendall $\tau_b$) & \OrdTauSmallR{} & & \anPost \\
 & Random halves of recording groups (Kendall $\tau_b$) & \OrdSplitSmallR{} & & \anCtl \\
\midrule
\stg{wmPurple}{\faBalanceScale}{Matched policy test} & Panel mean $-$ matched caches (RAD, matched-pressure stream) & \EpEOneComplete{} & \fpEpEOneComplete & \anReg \\
 & Panel mean $-$ matched caches (RAD, real stream) & \EpROneComplete{} & \fpEpROneComplete & \anReg \\
\bottomrule
\end{tabular}
\caption{\textbf{Held-out comparisons across the complete memory pipeline.} Positive differences favor the tested configuration; dots and bars show each row's estimate and interval on one difference scale, and each row is a separate comparison; the two rank-correlation rows are levels, not signed differences. RAD = Retention after Delivery; hit@3 is the share of reader probes with the correct next object among three guesses. Registered comparisons follow the pre-specified test families; ``Post-reg.'' indicates additional analyses, and ``Secondary'' identifies a held-out result not promoted as a pre-registered hypothesis. Interval procedures and all policy-level results are in the supplement.}
\label{tab:results}
\end{table*}

\subsection{Perception and admission: what enters the limited store}
\label{sec:perception}\label{sec:rq2}\label{sec:rq4}\label{sec:admission}
\begin{figure}[t]
\centering
\wmincludegraphics[width=0.82\columnwidth]{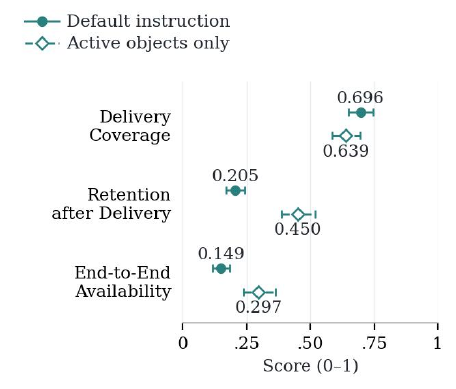}
\caption{\textbf{More selective name delivery improves small-memory retention.} FIFO with \ItemLimitSmall{} entries on the \HandSources{} held-out recordings with both instructions. The active-objects instruction reduces Delivery Coverage but increases Retention after Delivery and End-to-End Availability. Intervals use shared probe groups; this instruction experiment followed registration.}
\label{fig:perception}
\end{figure}

Before a replacement rule can help, perception must deliver the relevant object and admission must allow it into working memory. First, we test \emph{first-listed admission}: the memory admits only the first name reported for each frame, then applies its usual replacement rule. On the real stream, adding this rule to LRU raises RAD from \CapLvlLruSmallR{} to \CapLvlGatedLruSmallR{} with \ItemLimitSmall{} entries, a difference of \GateSmallR{}. The registered capacity comparison gives \GateInterR{}: this benefit is larger at \ItemLimitSmall{} than at \ItemLimitLarge{} entries. Admission and eviction therefore need separate evaluation, especially when names arrive faster than the small store can retain them.

A second intervention changes the upstream reports themselves. On \HandSources{} recordings, a post-registration paired instruction study asks the VLM for \emph{active objects} instead of all touchable objects. The input rate falls from \HandEpsDefaultC{} to \HandEpsHandledC{} names per second. Although Delivery Coverage changes from \HandDcDefaultC{} to \HandDcHandledC{}, FIFO's RAD increases by \HandGainSmall{} and its EtEA rises from \HandEteaDefaultSmall{} to \HandEteaHandledSmall{} (Fig.~\ref{fig:perception}). A rate-matched random-thinning control does not reproduce the RAD gain, so the gain cannot be explained by fewer arrivals alone. The two interventions operate at different boundaries: the instruction changes \emph{what perception reports}, while first-listed admission changes \emph{what working memory accepts}. Both can preserve relevant names before any replacement choice matters~\cite{cowan1999embedded,oberauer2002access}.

\subsection{Retention: what remains accessible as new names arrive}
\label{sec:mechanism}\label{sec:caches}\label{sec:capacity}\label{sec:rq1}\label{sec:separability}

The policy comparison asks a complementary question: once an identical stream has reached memory, what does the replacement rule contribute? \SigPassPanelCap{} of the \NumPanelPolicies{} cognitively inspired policies reproduce a source-predicted effect on synthetic event sequences; \SigAllControlsFail{} also demonstrate an effect absent from the five controls when the relevant operation is present. Behavioral tests establish whether the implemented effect occurs; separate held-out comparisons test whether it improves retention over caches that use the same ranking information: at \ItemLimitSmall{} entries, no panel policy exceeds its matched cache on either the matched-pressure or real stream. The real-stream panel-mean contrast is \EpROneComplete{}, and the matched-pressure contrast is \EpEOneComplete{}. Recency and frequency already contain information about future need~\cite{anderson1989human,salvatore2025emergent}. The result distinguishes the usefulness of a ranking signal from its added cognitive operation; it does not test the underlying theory of human memory.

\begin{figure*}[t]
\centering
\wmincludegraphics[width=0.78\textwidth]{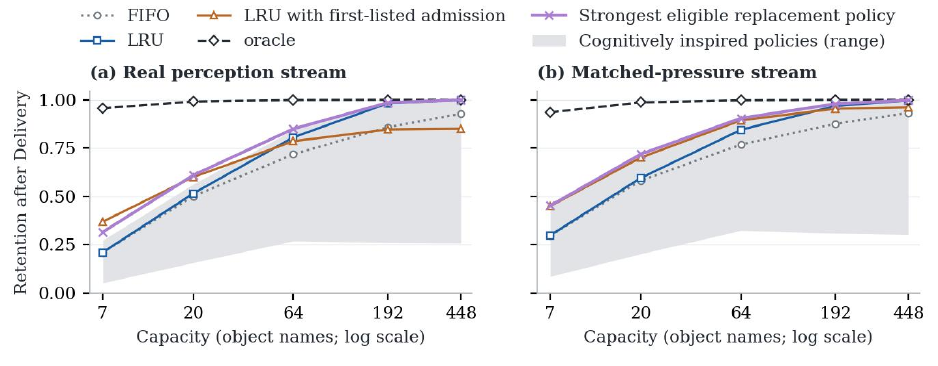}
\caption{\textbf{Capacity reveals different retention regimes.} Retention after Delivery on the real and matched-pressure streams, with the same future-action probes. FIFO, LRU, the first-name admission rule, the strongest development-selected eligible cache, a future-informed reference policy, and the range of cognitively inspired policy means are shown. The shaded range is \emph{not} an uncertainty band; per-policy curves and intervals are in the supplement.}
\label{fig:capacity}
\end{figure*}

Capacity changes the comparison itself (Fig.~\ref{fig:capacity}). At \ItemLimitSmall{} and \ItemLimitTwenty{} entries, FIFO and LRU retain similarly because the merge window prevents immediate repetitions; LRU becomes more effective at larger capacities when repeated names can be reused. Cache policies approach the future-informed reference as capacity grows, whereas policies that expire names before the store fills can stop benefiting from added entries. The figure therefore connects nominal capacity to actual use of the store rather than treating policy rankings as capacity-independent~\cite{denning1968working}. Complete curves, source-specific traces, distinct-name, and matched-byte controls appear in the supplement.

\subsection{Long-term retrieval: what can be recovered after eviction}
\label{sec:archive}

Working memory is not the assistant's entire past. We separately archive names that were evicted or never admitted, without imposing the working-memory capacity on that archive. At a probe, associative retrieval uses recently held names as cues and returns up to \ArcRetrieveK{} archived names, excluding recently reported names. This does not change RAD: it changes \emph{availability after retrieval}, the fraction of upcoming active objects available in the union of held and retrieved names. On the real stream, FIFO with \ItemLimitSmall{} entries rises from \ArcOffSmallR{} to \ArcOnSmallR{} after retrieval (\ArcGainSmallR{}), recovering \ArcRecoveredR{} of previously delivered target names no longer held. With the same maximum number of names presented to the reader, FIFO plus retrieval also exceeds FIFO with \WmFourteenNames{} working-memory entries by \GapEwMinusFourteenFifoR{}. This comparison demonstrates a complementary role for long-term storage, without treating archival capacity or retrieval as part of bounded working memory. The archive is queried by a bounded set of currently held cues; it is not given the future target and does not alter the seven-entry memory state. Matching the reader-list length against a larger working memory separates benefits of retrieval from merely presenting more names.

\subsection{Reader utility: whether the accessible list helps prediction}
\label{sec:usage}
\begin{figure}[t]
\centering
\wmincludegraphics[width=0.82\columnwidth]{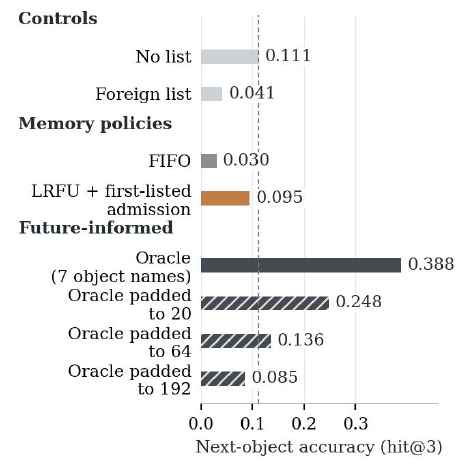}
\caption{\textbf{Relevant names help, while additional names dilute the signal.} Qwen3-VL top-three next-object accuracy (hit@3) on \STwoProbesNncf{} probes where the current object-name reply omits the target. The unrelated comparison list comes from another recording group. The future-informed reference uses later action annotations; padding leaves its target intact while adding names that are not upcoming targets.}
\label{fig:study2}
\end{figure}

Holding a name matters to an assistant only if its reader can use it. Qwen3-VL receives the current frame and a candidate memory list, then names up to three objects that could be used in the next action. We measure the share of probes at which the correct next object appears among these three guesses, called \emph{hit@3}. We evaluate \STwoProbes{} probes, including \STwoProbesNncf{} \emph{memory-dependent probes}, where the current object-name reply omits the target. This is an operational name for an opportunity to use memory: the current image may still contain other information, so a correct answer does not by itself prove memory use.

The registered list comparisons establish that \emph{which names are accessible} matters. A future-informed reference list containing the target achieves hit@3 \STwoOracleHitThree{}, compared with \STwoForeignHitThree{} for a list from another recording group (\STwoLOne{}); this reference cannot be deployed. A deployable ACT-R rule combining recency, frequency, and first-listed admission reaches \STwoWeightedNncf{}, compared with \STwoFifoNncf{} for FIFO (\STwoLTwo{}). The related LRFU policy with admission reaches \STwoGatedLrfuNncf{} (Fig.~\ref{fig:study2}). These compare \emph{memory lists}, not memory against the current frame; the target may still be visually inferable even when the current name reply omits it.

The registered padding experiment isolates list length from target availability. Starting with a seven-name reference list chosen using future actions, we add irrelevant names while retaining the target. Hit@3 falls from \STwoOracleHitThree{} to \RdPadMaxHit{} at \ItemLimitLarge{} entries (\RdPadMaxMinusOracle{} for padded minus short). Thus, retention, archival retrieval, and prompt composition should not be treated as one score: a larger available set need not yield a more useful reader input. The registered comparisons establish that useful list contents and list size both matter. All-probe no-list controls and post-registration rules that show the list only when the current name reply names no object further distinguish \emph{retaining} an object from \emph{choosing when to expose its name} (supplement). The latter comparisons are not used to claim a general improvement over a no-memory reader.

\subsection{Reproducibility across perception and event order}
\label{sec:order}\label{sec:rq3}

WM-Bench also tests the stability of policy comparisons. With \ItemLimitSmall{} entries on the real stream, \SepSepSmallR{} of the \SepPairsSmall{} pairs among \SepPoliciesSmall{} policies differ beyond the margin, and rankings from random halves of the recording groups agree (median Kendall $\tau_b=\SepTauSmallR{}$). On the \MolmoSources{} recordings available with a second perception VLM, Molmo2, the ordering of \TauVlmMolmoPolicies{} common policies agrees with Qwen3-VL (Kendall $\tau_b=\TauVlmMolmo{}$), despite the two VLMs reporting different names. A registered replay reverses the names within each frame without changing their timestamps; this tests the temporal-order assumptions relevant to serial-memory theories~\cite{murdock1962serial}. Reversal changes the advantage of Isolation-Weighted Decay, which favours a name preceded by a long time gap~\cite{brown2007temporal}, over FIFO by \RegRSeven{} (Table~\ref{tab:results}); a post-registration replay extends the test to the ranking of all policies. At \ItemLimitSmall{} entries on the real stream, the policy ranking has Kendall rank correlation~\cite{kendall1945treatment} $\tau_b=\OrdTauSmallR{}$ between listing and reversed order, compared with \OrdSplitSmallR{} between random halves of recording groups. Within-frame order is therefore part of the online interface that a reproducible benchmark must specify, not an implementation detail to leave implicit. Detailed order controls and the expanded capacity results appear in the supplement.

\paragraph{Reproducibility and release.}
The benchmark replays cached, timestamped name streams on CPU, keeps the independent scorer outside the policy interface, and retains held-out recording-group membership for paired inference. The registration record, policy specification cards, audit annotations, complete capacity curves, and reconstruction scripts accompany the release. This lets others test additional memory rules without repeating VLM decoding or changing the target population.

\section{Scope and Limitations}
\label{sec:scope}
WM-Bench isolates \emph{symbolic object-category memory}: it does not identify physical instances, store visual features, or measure a deployed assistant's complete latency. An audit of category matching estimates precision \AuditPrecision{}; instance-grounded and stricter-matching checks are in the supplement. Policies operate without task descriptions or attention cues, even though some cognitive sources assume them. The video/agent-memory proxies implement their symbolic replacement operations rather than their complete systems. Qwen3-VL supplies the main perception and reader tests; Molmo2 supports a partial cross-perception comparison. Finally, equal entry counts do not guarantee equal bytes or equal distinct categories; those controls are reported separately. The replay measures memory decisions, not full smart-glasses latency; action annotations inform the scorer only. Future instance-aware and task-conditioned assistants can reuse the stage-level scoring protocol.

\section{Conclusion}
\label{sec:conclusion}
WM-Bench evaluates useful object access across the memory pipeline of an egocentric assistant rather than reducing memory to final-answer accuracy. Replaying the same object-name streams through different policies lets us score perception, admission, bounded retention, retrieval, and reader use separately. The held-out results show that selective admission can improve retention in small memories, retrieval can recover older names, and short lists of relevant names help next-object prediction. Tests of predicted cognitive behavior and caches using the same ranking information clarify what each memory rule contributes. By exposing the reported names, held state, retrieved names, and reader response, WM-Bench supports reproducible comparisons and practical choices about which object names to keep accessible during a task.

\section*{Use of AI Assistants}
AI assistant revised some author-written passages and helped in figure and table presentation.


\end{document}